\documentclass[sigconf]{acmart}

\usepackage{enumitem}
\usepackage{multirow}
\usepackage{pifont}
\usepackage{colortbl}

\AtBeginDocument{%
  }

\setcopyright{acmlicensed}
\copyrightyear{2024}
\acmYear{2024}
\acmConference[CIKM '24] {Proceedings of the 33rd ACM International Conference on Information and Knowledge Management}{October 21--25, 2024}{Boise, ID, USA.}
\acmBooktitle{Proceedings of the 33rd ACM International Conference on Information and Knowledge Management (CIKM '24), October 21--25, 2024, Boise, ID, USA}
\acmDOI{10.1145/3627673.3679551}
\acmISBN{979-8-4007-0436-9/24/10}
\begin{document}

\title{DAMe: Personalized Federated Social Event Detection with Dual Aggregation Mechanism}

\author{Xiaoyan Yu}
\orcid{0009-0006-2314-7867}
\affiliation{%
  \institution{School of Computer Science and Technology, \\
  Beijing Institute of Technology}
  \city{Beijing}
  \country{China}}
\email{xiaoyan.yu@bit.edu.cn}

\author{Yifan Wei}
\orcid{0000-0002-8870-7304}
\affiliation{%
  \institution{School of Artificial Intelligence, \\
  University of Chinese Academy of Sciences}
  \city{Beijing}
  \country{China}}
\email{weiyifan2021@ia.ac.cn}

\author{Pu Li}
\orcid{0009-0007-1410-4096}
\affiliation{%
  \institution{
  Kunming University of Science and Techonology}
  \city{Kunming}
  \country{China}}
\email{lip@stu.kust.edu.cn}

\author{Shuaishuai Zhou}
\orcid{0009-0000-9464-1316}
\affiliation{%
  \institution{
  Kunming University of Science and Techonology}
  \city{Kunming}
  \country{China}}
\email{20222204121@kust.edu.cn}

\author{Hao Peng}
\orcid{0000-0003-0458-5977}
\authornote{Corresponding authors.}
\affiliation{%
  \institution{
  Beihang University}
  \city{Beijing}
  \country{China}}
\email{penghao@buaa.edu.cn}

\author{Li Sun}
\orcid{0000-0003-4562-2279}
\affiliation{%
  \institution{
  North China Electric Power University}
  \city{Beijing}
  \country{China}}
\email{ccesunli@ncepu.edu.cn}

\author{Liehuang Zhu}
\authornotemark[1]
\orcid{0000-0003-3277-3887}
\affiliation{%
  \institution{
  % School of Cyberspace Science and Technology, \\
  Beijing Institute of Technology}
  \city{Beijing}
  \country{China}}
\email{liehuangz@bit.edu.cn}

\author{Philip S. Yu}
\orcid{0000-0002-3491-5968}
\affiliation{%
  \institution{
  University of Illinois at Chicago}
  \city{Chicago}
  \state{IL}
  \country{USA}}
\email{psyu@uic.edu}

\renewcommand{\shortauthors}{Xiaoyan Yu et al.}

\begin{abstract}

Training social event detection models through federated learning (FedSED) aims to improve participants' performance on the task. 
However, existing federated learning paradigms are inadequate for achieving FedSED's objective and exhibit limitations in handling the inherent heterogeneity in social data.
This paper proposes a personalized federated learning framework with a dual aggregation mechanism for social event detection, namely DAMe.
We present a novel local aggregation strategy utilizing Bayesian optimization to incorporate global knowledge while retaining local characteristics.
Moreover, we introduce a global aggregation strategy to provide clients with maximum external knowledge of their preferences. 
In addition, we incorporate a global-local event-centric constraint to prevent local overfitting and ``client-drift''.
Experiments within a realistic simulation of a natural federated setting, utilizing six social event datasets spanning six languages and two social media platforms, along with an ablation study, have demonstrated the effectiveness of the proposed framework.
Further robustness analyses have shown that DAMe is resistant to injection attacks.
 
\end{abstract}


% \ccsdesc[300]{Computing methodologies~Artificial intelligence}




\maketitle

\section{Introduction}
    Social Event Detection (SED) aims to pinpoint unusual occurrences that involve specific times, locations, people, content, etc., in the real world from social media platforms \cite{peng2022reinforced}.
    Traditionally, individual platforms collect their own data to train SED models.
    However, users tend to post content across various platforms driven by personal preferences (e.g., linguistic preferences \cite{ren2021transferring} and social affiliations \cite{ren2022cross}).
    Consequently, the models trained individually by each platform are susceptible to their inherent biases, leading to a limited scope and incomplete detection of events.
    Meanwhile, due to privacy concerns, existing regulations prohibit organizations from sharing data without user consent \cite{sui2020feded}, making it unfeasible to centralize data for training purposes.
    In such scenarios, the most straightforward solution is implementing Federated Learning (FL) \cite{mcmahan2017communication}.
    In FL, each client (participant) trains a local model using their private data, while the central server facilitates information exchange among clients by iteratively aggregating the locally uploaded model weights.
    This paper initiates the study on Federated Social Event Detection (FedSED).

    Implementing SED through FL necessitates considering its inherent characteristics and challenges.
    \textbf{Firstly, FedSED aims to facilitate clients in achieving better performance on their respective data.}
    Unlike traditional FL \cite{mcmahan2017communication,li2020federated}, which prioritizes training a global model with optimal performance across all data, FedSED aims to facilitate information sharing among clients within a federated framework, thereby enhancing the performance of local models on respective data.
    This requires prioritizing client demands as the primary driving force throughout the federated process.
    \textbf{Secondly, social data sourced from various clients inhibit significant heterogeneity.}
    In practical FL scenarios \cite{li2020federated,huang2023federated}, data from different clients often exhibit non-independent and non-identically distributed characteristics (referred to as non-IID). 
    This non-IID nature of the data leads to clients converging in different directions, a phenomenon known as ``client-drift''. 
    While non-IID has been a longstanding issue in FL \cite{huang2021personalized}, it is further aggravated in the FedSED context since data from various platforms can differ in formats, languages, contents, etc.
    Consequently, addressing the challenges posed by non-IID social data, including multilingualism and multiplatform discrepancies, is paramount in FedSED.

    % Given FedSED's objective of enhancing local performance and inherent heterogeneity, personalized federated learning (pFL) approaches based on aggregation appear to be promising solutions.
    Given FedSED's objective of enhancing local performance and addressing inherent heterogeneity, personalized federated learning (pFL) approaches based on aggregation appear to be promising solutions.
    Firstly, in model-level aggregation \cite{luo2022adapt,zhang2020personalized}, 
        clients have access to the local models of all other clients and perform aggregation in their own preferred manner.
        However, this line of approaches is \textit{encumbered by substantial communication overhead and privacy concerns}, given that model parameters could be leveraged to recover private data \cite{zhu2019deep,zhao2020idlg}.
    Secondly, layer-level aggregation, \cite{sun2021partialfed}
        where clients learn strategies for selecting parameters from the global or local models as their respective local models. 
        Nevertheless, such a \textit{binary selection fails to capture the essential information and struggles to learn effective strategies that meet local objectives}. 
    Thirdly, parameter-level aggregation, 
        such as FedALA \cite{zhang2023fedala}, where clients learn aggregation weights for each parameter of the global and local models. 
        Nonetheless, attaining the optimal solution for all parameters proves to be highly challenging, if not practically unachievable. 
        Furthermore, the aggregation of the global model relies on the FedAvg strategy, which \textit{falls short in delivering the most advantageous knowledge to individual clients}.
    
    Expanding on the preceding discussion, we outline three crucial perspectives in developing a pFL framework for SED:
    % 1) On the server side, when dispatching the global model to each client, it is imperative to ensure that the model maximizes the information available for clients' personalization while reducing the heterogeneity issue it might bring;
    1) On the server side, when dispatching the global model to each client, it is imperative to strike a balance between maximizing the information available for client personalization and mitigating potential heterogeneity issues that may arise.
    2) On the client side, it is essential to retain a portion of local characteristics while integrating the knowledge provided by the server to prevent deviation from local objectives or local overfitting;
    3) In entirety, achieving a level of consensus between the global and local models on the representation of the same event is important to mitigate the impact of heterogeneity.
    
    In light of the abovementioned perspectives for developing the FedSED framework, this work proposes a novel \textbf{D}ual \textbf{A}ggregation \textbf{Me}chanism for Personalized Federated Social Event Detection, namely \underline{\textbf{\textit{DAMe}}}.
    The framework aims to enhance the performance of local models on local data through the collaborative efforts of the server and clients.
    DAMe comprises three components: local aggregation, global aggregation, and global-local alignment.
    For local aggregation, we employ Bayesian optimization to explore the ideal aggregation weight.
    This process facilitates the integration of extensive global knowledge while preserving the unique local characteristics to a great extent.
    For global aggregation, we construct a client graph on the server side and minimize the 2D structural entropy within it.
    Through this process, an optimal aggregation strategy is acquired to maximize the external knowledge available to each client while reducing global heterogeneity.
    For global-local alignment, we propose a global-local event-centric constraint to align the local event representation with the global event representation.
    % This ensures consistency between the information captured by the local models and the global understanding of specific events.
    This ensures that the local models acquire improved representations of social messages.
    We evaluate the proposed framework using six social event datasets covering six languages and two social media platforms.
    The experiments are conducted within a realistic simulation of a natural federated setting.
    Our experimental results and ablation study underscore the efficacy of DAMe for FedSED, with the potential to extend to other applications. 
    Further robustness analysis confirms that DAMe is resilient to federated injection attacks.

    In summary, the contributions of this work are as follows:
    \begin{itemize}[leftmargin=15pt]
        \item We pioneer the study on Federated Social Event Detection (FedSED) and present a pFL framework that satisfies the objective of the task.
        \item We propose a novel dual aggregation mechanism that maximizes the transfer of external knowledge from the server to the clients while enabling the clients to retain a portion of their own characteristics during local learning.
        \item We devise a global-local event-centric constraint to learn better message representation, meanwhile preventing local overfitting and ``client-drift''.
        % \item We devise a global-local event-centric constraint to mitigate data heterogeneity from multilingual and multiplatform sources.
        \item Extensive experiments conducted in natural federated settings have corroborated our proposed framework's effectiveness and demonstrated its robustness against federated injection attacks.
    \end{itemize}
\section{Related Work}

    \subsection{Social Event Detection}
        Social event detection, aiming to identify potential social events from social streams, is a longstanding and challenging task.
            Recent SED methods primarily rely on Graph Neural Networks (GNNs) \cite{peng2021streaming,peng2019fine,cao2021knowledge,cui2021mvgan,ren2021transferring,ren2022known, wei2023multi, ren2023uncertainty}.
            These approaches construct message graphs to represent social message data, integrating various attributes that complement each other and serve independent roles in propagating and aggregating text semantics.
            For instance, KPGNN \cite{cao2021knowledge} builds an event message graph using user, keyword, and entity attributes, then employs inductive Graph Attention Networks (GAT) to learn message representations. 
            Furthermore, some approaches adopt multi-view learning strategies to enhance the feature learning process. 
            MVGAN \cite{cui2021mvgan}, for instance, learns message features from both semantic and temporal views and incorporates an attention mechanism to fuse them. 
            ETGNN \cite{ren2023uncertainty} focuses on learning representations from co-hashtag, co-entity, and co-user views.
        However, current works have not yet explored methods utilizing federated learning to enhance the comprehensiveness and accuracy of SED.

    \subsection{Federated Learning}
        Federated Learning (FL), an advanced paradigm for decentralized data training, has garnered significant attention in recent years \cite{zhang2021survey}.
        FedAvg \cite{mcmahan2017communication} achieves collaborative learning across decentralized devices by locally training models and aggregating parameters on a central server.
        On that basis, FedProx \cite{li2020federated} introduces regularization that enhances model convergence and generalization by considering the smoothness of the global model in FL.
        
        Due to the statistical heterogeneity in FL, a centralized global model may lower the performance of certain participants \cite{zhang2023fedala}. 
        Hence, personalized FL has garnered significant attention \cite{tan2022towards}.
        We categorize pFL methods into three types.
        \textbf{Fine-Tuning-based Methods} involve learning a global model, which clients then fine-tune on their respective sides to achieve personalization \cite{fallah2020personalized,collins2021exploiting, yu2024neeko}.
        For instance, Per-FedAvg~\cite{fallah2020personalized} regards the global model as an initial shared model, allowing all clients to fine-tune it with local data to fit their specific needs.
        \textbf{Personalized Layer/Model-based Methods} offer flexibility in model architecture, allowing for variations such as sharing specific layers or training additional local models \cite{t2020personalized,li2021ditto,yang2023fedack,liu2023communication}.
        For example, FedACK \cite{yang2023fedack} employs GAN-based knowledge distillation for cross-model and cross-lingual social bot detection. 
        \textbf{Aggregation-based Methods} achieve personalization through server centrally aggregates specialized global models for the participants or clients directly exchange parameters among themselves in a decentralized setting, allowing them to select the information they desire \cite{huang2021personalized,li2021fedphp,zhang2020personalized,luo2022adapt,sun2021partialfed,chen2022personalized,zhang2023fedala}.
        
        Previous studies have overlooked the possibility of servers providing a suitable model while allowing clients to integrate helpful knowledge.
        To this end, this work focuses on personalization to enhance local performance via global and local aggregation.
\section{Preliminaries}\label{sec:pre}
    In this section, we outline the problem formulations of Federated Learning (FL) and Social Event Detection (SED), then define the threat model in the federated setting.

    \subsection{Federated Learning}\label{sec:setting}
        This paper considers FL with one central server and $K$ clients.
        % The dataset $\mathcal{D}_k = {\{x_i, y_i\}^{N_k}_{i=1}}$, locally collected by client $k$, remains inaccessible to others.
        The dataset $\mathcal{D}_k$, locally collected by client $k$, remains inaccessible to others.
        % Here, $x_i$ represents a data sample, $y_i$ is its corresponding label, and $N_k$ denotes the number of data samples owned by client $k$.
        Below is an overview of the training process for the classic FL algorithm FedAvg \cite{mcmahan2017communication}:
        \begin{description}[leftmargin=10pt]
            \item[\textbf{Step 1}] \textbf{Initialization.} 
            At the initial communication round $r=0$, all local model parameters of $K$ clients are initialized as the global model parameter: $\theta_0^{l_k} \leftarrow \theta_0^{g}$, where $\theta_0^{l_k}$ and $\theta_0^{g}$ denotes the model parameters of client $k$ and the server at round $0$, respectively.
            \item[\textbf{Step 2}] \textbf{Client Update.} 
            Each client $k$ trains the model on their private dataset $\mathcal{D}_k$ with task objective $\mathcal{L}(\mathcal{D}_k; \theta_r^{l_k})$.
            Then, upload the trained local model parameters $\theta_{r+1}^{l_k}$ to the server.
            \item[\textbf{Step 3}] \textbf{Server Execute.}
            The server aggregates the received parameters by $\theta_{r+1}^{g}= \sum_{k=1}^K \frac{N_k}{N_{sum}} \theta_{r+1}^{l_k}$, where $N_k$ denotes the number of data samples of client $k$, and $N_{sum}$ is the total number of data samples across all clients. 
            Then, the server distributes the new global model parameters to clients in the following round.
        \end{description}
        Iterate Steps 2 and 3 continuously until the final communication round.
        The global objective of the overall FL process is:
        \begin{equation}
            \underset{\theta}{\arg\min} \mathcal{L}(\theta) = \sum_{k=1}^{K} \mathcal{L}(\mathcal{D}_k; \theta).
        \end{equation}

    \subsection{Social Event Detection}
        % Given a collection of social messages $M$, the $i$-th message is represented as $m_i = \{d_i, t_i, a_i\}$, where $d_i$, $t_i$, and $a_i$ denote the text document, timestamp, and attributes (such as hashtags, users, entities) of $m_i$.
        % A social event detection algorithm aims to learn a model $f(M; \theta) = E$ from $M$, where $\theta$ is the model parameter, $E = \{e_j \mid 1 \leq j \leq |E|\}$ is the set of events (all labels).
        Given a collection of social messages $M$, a social event detection algorithm aims to learn a model $f(M; \theta) = E$ from $M$, where $\theta$ is the model parameter, $E = \{e_j \mid 1 \leq j \leq |E|\}$ is the set of events (all labels).

    \subsection{Threat Model}\label{sec:threat_model}
        Current federated paradigms are vulnerable to injection attacks \cite{lyu2022privacy}.
        In this work, the threat model is defined by the presence of malicious clients deliberately injecting backdoors within the training data (data poisoning attack) or uploading corrupted parameters to the server (model poisoning attack) to sabotage the FL process (e.g., performance, collaboration). 
        Such attacks could have profound repercussions on the global model and threaten FL's reliability and accuracy.
        We analyze the robustness of the proposed framework against injection attacks in Section \ref{sec:robust}.
% \newpage
\section{Methodology}

\begin{figure*}[!htbp]
        \centering
        \includegraphics[width=1.0\linewidth]{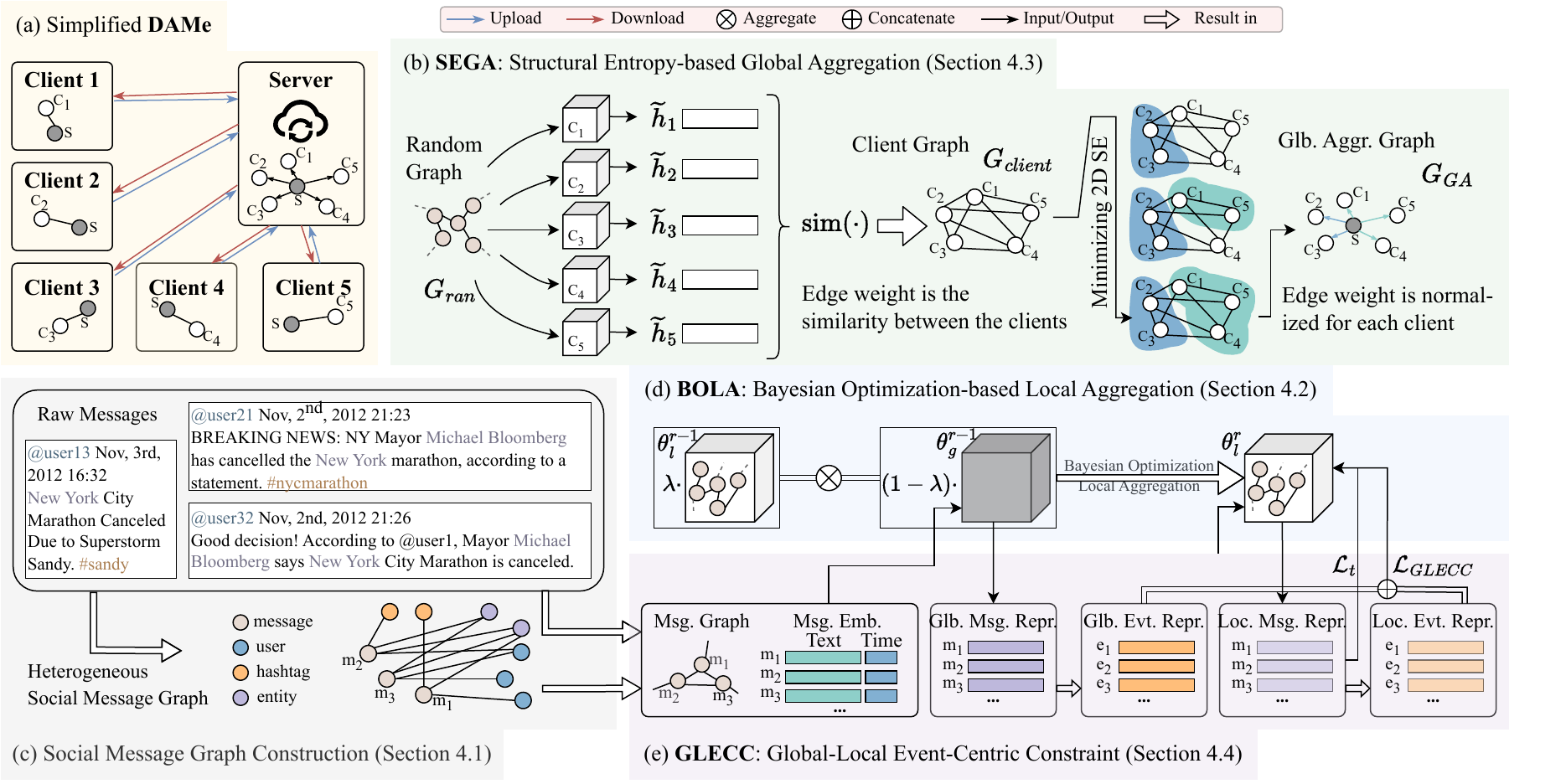}
        \caption{The overall framework of DAMe.}
        \label{fig:overall_frame}
    \end{figure*}

    This section presents the proposed framework, which consists of four key components: backbone model of SED, local aggregation, global aggregation, and global-local alignment.
    The overall framework is illustrated in Figure \ref{fig:overall_frame}.

    \subsection{Backbone Social Event Detection Model}\label{sec:backbone}

        For FedSED, we apply GAT (Graph Attention Network) as our backbone SED model and map the text encoding of different languages into a shared vector space.

        \paragraph{\textbf{Social Message Graph Construction}}
        As illustrated in Figure \ref{fig:overall_frame}(c),  attributes (including users, hashtags, and entities) are extracted from messages and connected with their corresponding messages, forming a heterogeneous social graph.
        Then, the heterogeneous social graph is projected onto a homogeneous social graph by retaining the original message nodes and adding edges connecting message nodes with shared attributes.
        In this graph, nodes represent messages, and edges signify the associations between messages. 

        \paragraph{\textbf{Social Message Representation}}
        The message embedding is obtained by concatenating the message's textual and temporal embedding.
        The temporal embedding corresponds to the message's timestamp in the OLE date format.
        Regarding textual embedding, accommodating the linguistic differences among clients is essential for FL. 
        Consequently, all clients utilize the pre-trained language model, SBERT-based (Sentence-BERT \cite{reimers2019sentence}) multilingual model \cite{reimers2020making} to encode the textual content of messages. 
        This implementation ensures that messages in diverse languages reside within a unified feature space.

    \subsection{Local Aggregation via Bayesian Optimization}\label{sec:local_aggr}

        We introduce a local aggregation mechanism, where clients learn a strategy that incorporates global knowledge while preserving their local characteristics rather than being directly overridden by the global model.
        The following optimization problem is formulated to describe the local aggregation process at the $r$-th communication round for client $k$:
        \begin{equation}
            \theta^{l_k}_{r+1} \leftarrow \widetilde{\theta}^{k}_r = \lambda_r^{k}\theta^{l_k}_r + (1-\lambda_r^{k})\theta^g_r,
        \end{equation}
        where $\theta^{l_k}_r$ and $\theta^g_r$ denote the local and global model parameters, respectively.
        $\lambda_r^{k} \in \mathbb{R}$ represents the aggregation weight (the weight of local preservation).
        Local aggregation strives to determine the optimal or near-optimal weight $\lambda_r$ that allows clients to acquire the maximum amount of knowledge.
        The process of $\theta^{l_k}_{r+1} \leftarrow \widetilde{\theta}^{k}_r$ is described in Section \ref{sec:loss}.
        
        Bayesian Optimization (BO) algorithm \cite{frazier2018tutorial} is a widely employed approach for optimizing functions with costly or challenging direct evaluations.
        We utilize \textbf{BO} for determining the aggregation weight $\lambda_r$ for \textbf{L}ocal \textbf{A}ggregation (\textbf{BOLA}), as shown in Figure \ref{fig:overall_frame}(d).
        BOLA is accomplished through a three-step BO procedure: 
            first defining the \textit{objective function}, 
            then modeling it using a \textit{Bayesian statistical model}, 
            and finally determining the subsequent sampling position by an \textit{acquisition function}.

        \subsubsection{Objective Function}
            The objective function for a single round of local aggregation can be formulated as follows (symbols denoting the $r$-th round and client $k$ are omitted for simplicity):
            \begin{equation}\label{equ:obj_func}
                \lambda = \underset{\lambda \in [\alpha,1]}{\arg \max} \, f(\lambda \cdot \theta^l + (1-\lambda) \cdot \theta^g, \mathcal{D}),
            \end{equation}
            suggesting that the aggregation weight $\lambda$ can be evaluated by observing the task-specific performance of the aggregated model on private datasets $\mathcal{D}$, e.g., NMI score for SED performance.
            $\lambda$ is controlled by a hyperparameter $\alpha \in [0, 1)$ to reduce search space.

        \subsubsection{Bayesian statistical model}
            To model the object function (Equation \ref{equ:obj_func}), Gaussian Process regression (GPR) is applied. 
            Specifically, for any finite set of points $\boldsymbol{\lambda} = [\lambda_1, \lambda_2,\ldots,\lambda_n]$, the joint distribution of the corresponding function values $f(\boldsymbol{\lambda}) = [f(\lambda_1), f(\lambda_2), \linebreak \ldots, f(\lambda_n)]$ follows a multivariate Gaussian distribution.
            Consequently, $f(\boldsymbol{\lambda})$ is characterized as a Gaussian process, denoted as:
            \begin{equation}
                f\left(\boldsymbol{\lambda}\right) \sim \operatorname{Normal}\left(\boldsymbol{\mu}\left(\boldsymbol{\lambda}\right), \boldsymbol{\kappa}\left(\boldsymbol{\lambda}, \boldsymbol{\lambda}\right)\right),
            \end{equation}
            where $\boldsymbol{\mu}(\cdot)$ and $\boldsymbol{\kappa}(\cdot)$ denote the mean and kernel functions, respectively. 
            The learnable parameters in these functions can be estimated through maximum likelihood estimation.

            Applying Bayes’ rule, we obtain a joint probability distribution:
            \begin{equation}
                \left[\begin{array}{c}
                f(\boldsymbol{\lambda}) \\
                f(\boldsymbol{\lambda}^*)
                \end{array}\right] \sim \operatorname{Normal} \left(\left[\begin{array}{l}
                \boldsymbol{\mu}({\boldsymbol{\lambda}}) \\
                \boldsymbol{\mu}({\boldsymbol{\lambda}^*})
                \end{array}\right],\left[\begin{array}{ll}
                \mathcal{K} & \mathcal{K}_{*} \\
                \mathcal{K}_{*}^T & \mathcal{K}_{**}
                \end{array}\right]\right),
            \end{equation}
            where $\lambda^*$ denotes the current optimal value of $\lambda$, which serves as the objective of the optimization process.
            $\mathcal{K} = \boldsymbol{\kappa}(\boldsymbol{\lambda}, \boldsymbol{\lambda})$, $\mathcal{K}_{*} = \boldsymbol{\kappa}(\boldsymbol{\lambda}, \boldsymbol{\lambda}^*)$, and $\mathcal{K}_{**} = \boldsymbol{\kappa}(\boldsymbol{\lambda}^*, \boldsymbol{\lambda}^*)$.
            Based on the joint distribution of $f(\boldsymbol{\lambda})$ and $f(\boldsymbol{\lambda}^*)$, the conditional distribution is as follows:
            \begin{equation}\label{equ:conditional}
                \begin{aligned}
                    f(\boldsymbol{\lambda}^*) \mid f(\boldsymbol{\lambda} )  \sim &\operatorname{Normal}  (\boldsymbol{\mu}_*, \boldsymbol{\kappa}_*), \\
                    \boldsymbol{\mu}_* = \boldsymbol{\mu}(\boldsymbol{\lambda}_*) +\mathcal{K}_*^T& \mathcal{K}^{-1}(f(\boldsymbol{\lambda})-\boldsymbol{\mu}(\boldsymbol{\lambda})), \\
                    \boldsymbol{\kappa}_* =\mathcal{K}_{**}-& \mathcal{K}_*^T \mathcal{K}^{-1} \mathcal{K}_*.
                \end{aligned}
            \end{equation}
            Through the above equations, it is observed that the posterior distribution's statistical properties $\boldsymbol{\mu}_*$ and  $\boldsymbol{\kappa}_*$ are modeled using GPR on the prior distribution's mean function $\boldsymbol{\mu}(\boldsymbol{\lambda})$ and covariance function $\boldsymbol{\kappa}\left(\boldsymbol{\lambda}, \boldsymbol{\lambda}\right)$.

        \subsubsection{Acquisition function}
            
            The acquisition function is used to determine the next aggregation weight.
            In this work, we apply the Expected Improvement (EI) \cite{mockus1974bayesian,jones1998efficient} criterion and the Upper Confidence Bound (UCB) \cite{srinivas2010gaussian} as acquisition functions.
            
            \paragraph{\textbf{Expected Improvement (EI)}}
                The calculation of the objective function ${f}(\boldsymbol{\lambda}^* \mid \theta^l, \theta^g, \mathcal{D})$ necessitates processing the entire dataset $\mathcal{D}$, which makes obtaining the next weight's objective function value costly.
                Therefore, we calculate the expected improvement value of the next weight with the aim of maximizing it.
                EI is computed as:
                \begin{equation}
                    \lambda_{t} = \underset{\lambda \in \boldsymbol{\lambda}}{\arg\max} \mathbb{E}_{f(\lambda) \sim \mathcal{N}
                    \left(\boldsymbol{\mu_{t-1}}\left(\lambda\right), \boldsymbol{\kappa}_{t-1}\left(\lambda,\lambda\right)\right)}\left[\max \left(f(\lambda)-f_{t-1}^{*}, 0\right)\right],
                \end{equation}
                where $\mathbb{E}[\cdot]$ denotes the expectation computed under the posterior distribution (Equation \ref{equ:conditional}).
                $\lambda_{t}$ denotes the optimal $\lambda$ at the $t$-th step, and $f_{t-1}^{*}$ denotes the optimal result during the first $t-1$ iterations.
                
            \paragraph{\textbf{Upper Confidence Bound (UCB)}}
                The UCB algorithm chooses the weight with the highest upper confidence bound for exploration, aiming to converge towards weights with higher actual reward values.
                It is defined as:
                \begin{equation}
                    \lambda_{t}= \underset{\lambda \in \boldsymbol{\lambda}}{\arg\max} \mu_{t-1}(\lambda)+\beta_t^{\frac{1}{2}} \sigma_{t-1}(\lambda).
                \end{equation}
                Here, $\beta_t > 0$ is a learnable parameter derived from theoretical analysis that increases over time. 
                $\sigma(\cdot)$ denotes the standard deviation function.
             
                Given the intricate and non-convex nature of the objective function $f(\cdot)$ \cite{hoffman2011portfolio}, we employ a mixed acquisition strategy of incorporating EI and UCB.

    \subsection{Global Aggregation via 2D Structural Entropy Minimization}\label{sec:global_aggr}
        
        Under the federated framework described in Section \ref{sec:setting}, personalized global aggregation aims to provide clients with maximum external information by producing global models that can benefit individual clients more.
        The server needs an aggregation strategy that considers client heterogeneity and individual characteristics to maximize external knowledge for all clients.
        To achieve this objective, we construct a client graph $G_{client}$ based on clients' similarity. 
        By minimizing the two-dimensional \textbf{S}tructural \textbf{E}ntropy (2DSE) of $G_{client}$, a graph capturing the internal similarities among clients is obtained, finalizing the \textbf{G}lobal \textbf{A}ggregation strategy for each client (\textbf{SEGA}).
        This process is demonstrated in Figure \ref{fig:overall_frame}(b).

        $G_{client}$ is an undirected, fully connected, weighted graph consisting of $K$ nodes corresponding to $K$ clients, with their similarities as edge weights.
        The similarity between client models can be estimated by providing them with the same input and measuring the similarity between their respective outputs.
        On this basis, the server first generates a random graph $G_{random}$ as input to all client models \cite{holland1983stochastic}.
        With graph pooling \cite{lee2019self}, the server obtains different client models' representations of the same graph, and the similarity between client $u$ and $v$ is calculated as:
        \begin{equation}
            \text{sim}(u, v) = \frac{\tilde{h}_u \cdot \tilde{h}_v}{\|\tilde{h}_u\| \|\tilde{h}_v\|},
        \end{equation}
        where $\tilde{h}_u$ is the averaged output of all node embeddings in the input graph $G_{random}$ and $\text{sim}(u, u)=1$.
    
        Upon constructing the client graph $G_{client}=(V,E)$, we minimize the 2DSE of the graph, resulting in a partitioned graph, which serves as the basis for the aggregation strategy.
        Suppose $\mathcal{P} = \{X_1, X_2, \dots, X_L\}$ forms partitions of nodes in $V$, where $L\leq K$ is the total number of partitions, $V=\{c_1, ..., c_K\}$ represents the set of client nodes, and $X_l$ denotes the $l$-th partition that contains specific client node(s).
        The 2DSE of the client graph $G_{client}$ is calculated as follows \cite{cao2024hierarchical,li2016structural}:
        \begin{equation}
            \begin{aligned} 
            \mathcal{H}^\mathcal{P}(G_{client})
            =&\sum_{l=1}^L \mathcal{H}_{X_l}^\mathcal{P}(G_{client})\\
            % = & \sum_{l=1}^L \frac{\operatorname{vol}(X_l)}{\operatorname{vol}(G_{client})} \cdot H\left(\frac{d_1^{(l)}}{\operatorname{vol}(X_l)}, \ldots, \frac{d_{n_l}^{(l)}}{\operatorname{vol}(X_l)}\right)\\ 
            % & -\sum_{l=1}^L \frac{g_l}{\operatorname{vol}(G_{client})} \log _2 \frac{\operatorname{vol}(X_l)}{\operatorname{vol}(G_{client})} \\ 
            = & -\sum_{l=1}^L \frac{\operatorname{vol}(X_l)}{\operatorname{vol}(G_{client})} \sum_{i=1}^{n_l} \frac{d_i^{(l)}}{\operatorname{vol}(X_l)} \log _2 \frac{d_i^{(l)}}{\operatorname{vol}(X_l)} \\
            & - \sum_{l=1}^L\frac{g_l}{\operatorname{vol}(G_{client})} \log _2 \frac{\operatorname{vol}(X_l)}{\operatorname{vol}(G_{client})},
            \end{aligned}
        \label{equ:2dse}
        \end{equation}
        where $L$ denotes the number of total partitions, $n_l$ denotes the number of client nodes in partition $X_l$, $d_i^{(l)}$ denotes the degree of the $i$-th client node in $X_l$, $\operatorname{vol}(\cdot)$ computes the volume, and $g_l$ denotes the sum of degrees of edges with one endpoint in partition $X_l$.
        The objective of the minimization process is to assign each client node $c_k$ to a distinct partition $X_l$.
        Specifically, each client node is initially treated as an individual partition. 
        New partitions are formed by iteratively merging different partitions. 
        The changes in the 2DSE before and after merging are observed to identify the partitioning scheme that yields the lowest overall 2DSE and generates the desired partitions.
        We leverage the greedy strategy in \cite{li2016structural} to minimize 2DSE.
        The difference in 2DSE before and after merging $X_i$ and $X_j$ into $X_l$ is calculated as follows:
        \begin{equation}
            \begin{aligned} 
                \Delta SE= & SE_{new}-SE_{old} \\
                         = & H^{\mathcal{P}'}(G_{client})- H^{\mathcal{P}}(G_{client})\\
                        = & H_{X_l}^{\mathcal{P}'}(G_{client}) - H_{X_i}^\mathcal{P}(G_{client}) - H_{X_j}^\mathcal{P}(G_{client}),\\
            \end{aligned}
        \end{equation}
        where the calculation of $H_{X_l}^{\mathcal{P}'}(G_{client})$, $H_{X_i}^\mathcal{P}(G_{client})$, and $H_{X_j}^\mathcal{P}(G_{client})$ follows Equation \ref{equ:2dse}, to compute the 2DSE of partition $X_*$ under the partion $\mathcal{P}^*$. 
        $H^{\mathcal{P}'}(G_{client})$ denotes the 2DSE of $G_{client}$ obtained by merging $X_i$ and $X_j$ into $X_l$.
        Note that we always merge the two partitions with the smallest $\Delta$SE until all $\Delta$SE $\geq 0$, thus obtaining the final partitions $\mathcal{P}_{final}=\{X_1,...,X_L\}$.
        Based on the final partition, the global aggregation strategy aims to aggregate information within each partition. 
        Specifically, in the $j$-th partition $X_j$, all client nodes are connected by edges weighted according to their similarities. 
        For all nodes in the partition, the global model $\theta_u^g$ for client $u$ is obtained by:
        \begin{equation}
            \theta_u^g = \sum_{v \in N(u)} \alpha_{uv} \cdot \theta_{v}^l,
        \end{equation}
        where $v \in N(u)$ represents the node within the same partition as $u$, $\theta_{v}^l$ is the local model of client $v$, and $\alpha_{uv}$ is the normalized weight between client $u$ and client $v$, computed as:
        \begin{equation}
            \alpha_{uv} = \frac{\exp(\text{sim}(u,v))}{\sum_{v \in N(u)} \exp(\text{sim}(u,v))}.
        \end{equation}

    \subsection{Local Optimization}\label{sec:loss}

        In Section \ref{sec:local_aggr}, we introduced a local aggregation strategy that aggregates $\theta^g_r$ and $\theta^l_r$ into $\widetilde{\theta}^l_r$. 
        This section describes the local optimization of $f(\widetilde{\theta}^l_r)$ with local data, maintaining the proximity between the local and the global models while preventing overfitting to the local data.
        The overall process is shown in Figure \ref{fig:overall_frame}(e).

        \paragraph{\textbf{Step 1: Triplet Loss}}
        Essentially, the objective of SED is to maximize similarity among messages belonging to the same event.
        Current approaches in SED \cite{cao2021knowledge} employ contrastive triplet loss to guide the optimization process.
        The triplet loss is computed as:
        \begin{equation}
            \begin{split}
                \mathcal{L}_t^* = \sum_{\left(m_i, m_i+, m_i-\right) \in \{T\}} \max\{D\left(h_{m_i}^*, h_{m_i+}^*\right) -     \\
                D\left(h_{m_i}^*, h_{m_i-}^*\right) + a, 0\},
            \end{split}
            \label{equ:trip_ls}
        \end{equation}
        where $\left(m_i, m_i+, m_i-\right) \in \{T\}$ is a set of constructed triples, $m_i$ represents the anchor sample, $m_i+$ represents the positive sample (i.e., a sample from the same class as the anchor sample), and $m_i-$ represents the negative sample (i.e., a sample from a different class than the anchor sample).
        $D(\cdot)$ calculates the Euclidean distance between samples, $h_{m_i}$ denotes the representation of message $m_i$, $a$ is the margin parameter, and $* = \{g, l\}$ denotes global or local.

        \paragraph{\textbf{Step 2: Global-Local Event-Centric Constraint}}
        In FedSED, data among clients exhibits non-IID characteristics. 
        This non-IID nature leads to ``client-drift'', resulting in low model utility.
        Building on this observation, our study introduces an \textbf{E}vent-\textbf{C}entric \textbf{C}onstraint that aligns the \textbf{G}lobal and \textbf{L}ocal models closer (\textbf{GLECC}).
        Firstly, the client obtains message representations from global model $f(\theta^g_r)$ and aggregated model $f(\widetilde{\theta}^l_r)$.
        Then, we learn the event representation from $f(\theta^g_r)$ and $f(\widetilde{\theta}^l_r)$ based on the representations of the messages within each event:
        \begin{equation}
            \boldsymbol{h_{e_i}^*}=\frac{1}{N_{e_i}}\sum_{j=1}^{N_{e_i}} \{\boldsymbol{h_{m_j}^*} | \forall m_j \in e_i\},
        \end{equation}
        where $N_{e_i}$ is the total number of messages in event $e_i$, and $* = \{g, l\}$ denotes global or local.
        Finally, GLECC is calculated with pairwise loss \cite{ren2022known} as:
        \begin{equation}
                \mathcal{L}_{GLECC} = \frac{1}{N_E}\sum_{i=1}^{N_E} D\left(h_{e_i}^g, h_{e_{i}}^l\right),
        \end{equation}
        where $N_E$ denotes the number of events in the current batch.
        By pulling closer the representations of the same event in both global and local models, the server and client establish a mutual consensus on representation learning.
        This alignment mitigates the risk of overfitting to local data, preventing divergence from the global context, as well as the tendency to solely pursue the global objective while disregarding local characteristics.
        
        \paragraph{\textbf{Step 3: Overall Loss}}
        The overall loss during the optimization process is calculated as follows:
        \begin{equation}
            \mathcal{L} = \mathcal{L}_t^l + \alpha \mathcal{L}_{GLECC},
        \end{equation}
        where $\alpha$ is calculated as:
        \begin{equation}
        \alpha = \operatorname{exp}(\operatorname{min}\{(\mathcal{L}_t^l - \mathcal{L}_t^g),0\})
            % \alpha = 
            % \begin{cases}
            %     e^{-(\mathcal{L}_t^g - \mathcal{L}_t^l)} & \text{if } \mathcal{L}_t^g > \mathcal{L}_t^l, \\ 
            %     1 & \text{if } \mathcal{L}_t^g \leq \mathcal{L}_t^l,
            % \end{cases}
        \end{equation}
        suggesting that the local model should only learn from the global model when the global model demonstrates better performance on local data.
        Note that, the global model $f(\theta^g_r)$ remains fixed throughout the whole process, training $f(\widetilde{\theta}^l_r)$ into $f(\theta^l_{r+1})$.
\section{Experimental Setups}
    This section introduces the experiment setups.
    We outline the following research questions (RQs) as guidelines for our experiments:
    
    \begin{itemize}[leftmargin=15pt]
        \item \textbf{RQ1}: Compared to existing FL approaches, can the proposed DAMe improve local performance?
        \item \textbf{RQ2}: Is the proposed framework robust (able to withstand injection attacks) in the setting of federated learning?
        \item \textbf{RQ3}: How does each component of the proposed framework contribute to the overall performance?
        \item \textbf{RQ4}: Regarding computation and communication, is the proposed framework efficient?
    \end{itemize}

    \subsection{Datasets}
        We conducted experiments on 6 datasets, spanning 6 languages and 2 platforms.
        Table \ref{tab:statisitcs} presents the statistics of all datasets.
        The English Twitter \cite{mcminn2013building} dataset, the French Twitter \cite{mazoyer2020french} dataset, and the Arabic Twitter \cite{alharbi2021kawarith} dataset are publicly available.
        We collect the rest of the datasets from Weibo in Chinese and Twitter in Japanese and German.
        First, we extracted key events from Wikipedia\footnote{\url{https://en.wikipedia.org/wiki/Portal:Current_events}} pages in multiple languages for 2018. 
        Subsequently, we retrieved relevant posts from Twitter or Weibo using event keywords and crawled them to construct the datasets. 
        The events listed on Wikipedia pages in different languages are customized according to users' preferences in that language, making the obtained dataset closely resemble real-world distributions.
        % We made the datasets publicly accessible on GitHub\footnote{\url{https://github.com/XiaoyanWork/pFedSED}}.

    \subsection{Federated Setting}
        Prior works \cite{liu2023communication,yang2023fedack} often partition a single dataset into multiple clients to mimic the federated setting.
        However, such practices fail to capture the non-IID nature of real-world data. 
        This study breaks this cycle by treating each dataset as an independent client in the experiments.
        This enables us to replicate the complexities of real-world data distribution across platforms more accurately. 
        By preserving the inherent non-IID characteristics of the data, we aim to enhance the fidelity of our federated learning experiments and provide insights that are more applicable to practical scenarios.
        In our experiments, we utilize a setup consisting of one server and six clients, each with social message data in distinct languages.

        \begin{table}[t]
        \caption{Statistics of the datasets.}
        \centering
        \resizebox{0.95\linewidth}{!}{
            \begin{tabular}{lccc}
            \toprule
            Dataset          & \# Messages   & \# Events & Avg. Length      \\
            \hline\hline
            English Twitter  & 68,841        & 503       & 27.03            \\
            French Twitter   & 64,516        & 257       & 51.77            \\
            Arabic Twitter   & 9,070         & 7         & 53.18            \\
            Japanese Twitter & 60,530        & 189       & 38.66            \\
            German Twitter   & 90,091        & 179       & 36.76            \\
            Chinese Weibo    & 71,846        & 221       & 30.60            \\
            \bottomrule
            \end{tabular}}
        \label{tab:statisitcs}
        \end{table}
    
    \subsection{Baselines}
% 是否需要加一个实验：KPGNN as backbone 来验证SBERT同一空间的意义
        In addition to \textbf{Local} training without parameter sharing, we compare DAMe with two categories of FL methods in the task of SED:
        (1) \textit{Classic FL methods:} 
            \textbf{FedAvg} \cite{mcmahan2017communication} aggregates a weighted global model for all clients. 
            \textbf{FedProx} \cite{li2020federated} introduce regularization to alleviate disparities between the global and local models.
        (2) \textit{Personalized FL methods: }
            \textbf{Per-FedAvg} \cite{fallah2020personalized} fine-tunes the global model on the client side to achieve personalization. 
            In \textbf{Ditto} \cite{li2021ditto}, each client learns an additional personalized model by incorporating a proximal term to extract information from the updated global model. 
            \textbf{SFL} \cite{chen2022personalized} constructs a client graph on the server and aggregates personalized models for each client via structural information. 
            % \textbf{APPLE} \cite{luo2022adapt} locally aggregates client models by learning the weights in each training batch rather than solely during local initialization.
            In \textbf{APPLE} \cite{luo2022adapt}, clients have access to all other clients' models and aggregates locally.
            \textbf{FedALA} \cite{zhang2023fedala} dynamically aggregates the global and local parameters at a fine-grained level based on the local objective.

        \begin{table*}[htbp]
        \aboverulesep=0ex
        \belowrulesep=0ex
        \caption{Results for DAMe (in \colorbox[HTML]{FFCC99}{orange} background) comparing with all baseline methods. The best result is marked in \textbf{bold}.} 
        \centering
        \resizebox{1.0\linewidth}{!}{
        \begin{tabular}{l|ccc|ccc|ccc|ccc}
        \toprule
        Clients    & \multicolumn{3}{|c|}{English Twitter}         & \multicolumn{3}{|c|}{French Twitter}          & \multicolumn{3}{|c|}{Arabic Twitter}          & \multicolumn{3}{|c}{Japanese Twitter}              \\
        \hline
        Metrics    & NMI           & AMI           & ARI           & NMI           & AMI           & ARI           & NMI           & AMI           & ARI           & NMI           & AMI           & ARI                \\
        \hline
        Local      & .69 $\pm$ .00 & .52 $\pm$ .00 & .13 $\pm$ .00 & .69 $\pm$ .00 & .62 $\pm$ .00 & .15 $\pm$ .00 & .55 $\pm$ .00 & .55 $\pm$ .00 & .50 $\pm$ .00 & .60 $\pm$ .00 & .50 $\pm$ .00 & .11 $\pm$ .01      \\
        \hline
        FedAvg     & .73 $\pm$ .00 & .58 $\pm$ .00 & .20 $\pm$ .01 & .74 $\pm$ .00 & .68 $\pm$ .00 & .21 $\pm$ .01 & .64 $\pm$ .00 & .64 $\pm$ .00 & .58 $\pm$ .00 & .64 $\pm$ .00 & .56 $\pm$ .00 & .18 $\pm$ .01      \\
        FedProx    & .71 $\pm$ .01& .55 $\pm$ .01& .16 $\pm$ .02& .70 $\pm$ .01& .64 $\pm$ .01& .17 $\pm$ .00 & .59 $\pm$ .01 & .59 $\pm$ .01 & .54 $\pm$ .02 & .62 $\pm$ .00 & .53 $\pm$ .00 & .14 $\pm$ .01      \\
        \hline
        Per-FedAvg & .72 $\pm$ .00 & .56 $\pm$ .00 & .17 $\pm$ .01 & .70 $\pm$ .00 & .64 $\pm$ .00 & .18 $\pm$ .01 & .62 $\pm$ .01 & .62 $\pm$ .00 & .57 $\pm$ .00 & .62 $\pm$ .00 & .53 $\pm$ .00 & .15 $\pm$ .01      \\
        Ditto      & .73 $\pm$ .00 & .58 $\pm$ .00 & .20 $\pm$ .01 & .73 $\pm$ .00 & .68 $\pm$ .00 & .21 $\pm$ .01 & .63 $\pm$ .02 & .62 $\pm$ .02 & .55 $\pm$ .02 & .64 $\pm$ .00 & .56 $\pm$ .00 & .18 $\pm$ .01      \\
        SFL        & .69 $\pm$ .00 & .52 $\pm$ .00 & .12 $\pm$ .00 & .69 $\pm$ .00 & .62 $\pm$ .00 & .15 $\pm$ .00 & .56 $\pm$ .00 & .56 $\pm$ .00 & .50 $\pm$ .01 & .60 $\pm$ .00 & .50 $\pm$ .00 & .11 $\pm$ .00      \\
        APPLE      & .71 $\pm$ .00 & .55 $\pm$ .00 & .16 $\pm$ .01 & .70 $\pm$ .00 & .63 $\pm$ .00 & .16 $\pm$ .00 & .70 $\pm$ .02& .69 $\pm$ .01& .73 $\pm$ .01& .61 $\pm$ .01& .52 $\pm$ .03& .12 $\pm$ .00      \\
        FedALA     & .74 $\pm$ .00 & .59 $\pm$ .00 & .21 $\pm$ .00 & .74 $\pm$ .00 & .69 $\pm$ .00 & .22 $\pm$ .00 & .63 $\pm$ .00 & .63 $\pm$ .00 & .57 $\pm$ .00 & .65 $\pm$ .00 & .56 $\pm$ .00 & .20 $\pm$ .01      \\ 
        \hline
        \rowcolor[HTML]{FFCC99}
        DAMe (Ours)       & \textbf{.75 $\pm$ .00} & \textbf{.61 $\pm$ .00} & \textbf{.21 $\pm$ .00} & \textbf{.76 $\pm$ .00} & \textbf{.70 $\pm$ .00} & \textbf{.23 $\pm$ .02} & \textbf{.86 $\pm$ .00} & \textbf{.86 $\pm$ .00} & \textbf{.89 $\pm$ .00} & \textbf{.67 $\pm$ .00} & \textbf{.59 $\pm$ .00} & \textbf{.25 $\pm$ .01}      \\ 
        % \rowcolor[HTML]{FFCC99}
        promotion  & $\uparrow$.01 & $\uparrow$.02 & $\uparrow$.01 & $\uparrow$.02 & $\uparrow$.01 & $\uparrow$.01 & $\uparrow$.16 & $\uparrow$.17 & $\uparrow$.16 & $\uparrow$.02 & $\uparrow$.03 & $\uparrow$.05      \\
        \hline\hline
        Clients    & \multicolumn{3}{|c|}{German Twitter}          & \multicolumn{3}{|c|}{Chinese Weibo}           & \multicolumn{3}{|c|}{Average on all clients}                 & \multicolumn{3}{|c}{Avg. Gain (compare to Local)}  \\
        \hline
        Metrics    & NMI           & AMI           & ARI           & NMI           & AMI           & ARI           & NMI           & AMI           & ARI           & NMI           & AMI           & ARI                \\
        \hline
        Local      & .41 $\pm$ .00 & .32 $\pm$ .00 & .04 $\pm$ .00 & .47 $\pm$ .00 & .34 $\pm$ .00 & .05 $\pm$ .00 & .57           & .48           & .16           & -             & -             & -                  \\
        \hline
        FedAvg     & .46 $\pm$ .00 & .39 $\pm$ .01 & .08 $\pm$ .01 & .53 $\pm$ .00 & .43 $\pm$ .00 & .11 $\pm$ .00 & .62           & .55           & .23           & .04           & .06           & .07                \\
        FedProx    & .43 $\pm$ .00 & .35 $\pm$ .00 & .06 $\pm$ .00 & .48 $\pm$ .01& .37 $\pm$ .00 & .07 $\pm$ .00 & .59           & .51           & .19           & .02           & .03           & .03                \\
        \hline
        Per-FedAvg & .44 $\pm$ .00 & .36 $\pm$ .00 & .06 $\pm$ .00 & .49 $\pm$ .00 & .38 $\pm$ .00 & .08 $\pm$ .01 & .60           & .51           & .20           & .02           & .03           & .04                \\
        Ditto      & .46 $\pm$ .00 & .39 $\pm$ .00 & .08 $\pm$ .00 & .53 $\pm$ .00 & .43 $\pm$ .00 & .11 $\pm$ .01 & .62           & .54           & .22           & .04           & .06           & .07                \\
        SFL        & .41 $\pm$ .00 & .32 $\pm$ .00 & .04 $\pm$ .00 & .47 $\pm$ .00 & .35 $\pm$ .00 & .05 $\pm$ .00 & .57           & .48           & .16           & .00           & .00           & .00                \\
        APPLE      & .43 $\pm$ .01& .35 $\pm$ .01& .06 $\pm$ .00& .48 $\pm$ .00& .36 $\pm$ .00 & .06 $\pm$ .00 & .61           & .52           & .22           & .01           & .02           & .01                \\
        FedALA     & .46 $\pm$ .00 & .39 $\pm$ .00 & .08 $\pm$ .00 & .53 $\pm$ .00 & .43 $\pm$ .00 & .12 $\pm$ .01 & .63           & .55           & .23           & .05           & .06           & .09                \\ 
        \hline
        \rowcolor[HTML]{FFCC99}
        DAMe (Ours)       & \textbf{.48 $\pm$ .00} & \textbf{.40 $\pm$ .00} & \textbf{.09 $\pm$ .00} & \textbf{.56 $\pm$ .00} & \textbf{.46 $\pm$ .00} & \textbf{.16 $\pm$ .01} & \textbf{.68}           & \textbf{.60}           & \textbf{.30}           & \textbf{.07}           & \textbf{.09}           & \textbf{.14}                \\ 
        % \rowcolor[HTML]{FFCC99}
        promotion  & $\uparrow$.02 & $\uparrow$.01 & $\uparrow$.01 & $\uparrow$.03 & $\uparrow$.03 & $\uparrow$.04 & $\uparrow$.05 & $\uparrow$.05 & $\uparrow$.07 & $\uparrow$.02 & $\uparrow$.03 & $\uparrow$.05      \\
        \bottomrule
        \end{tabular}
        }
        \label{tab:rq1}
        \end{table*}

    \subsection{Implementation Details}
        The experiments are implemented using the PyTorch framework and run on a machine with eight NVIDIA Tesla A100 (40G) GPUs.
        We randomly sample 70\%, 20\%, and 10\% for training, testing, and validation as common studies on SED \cite{cao2021knowledge,ren2022known}.
        For all methods, we employ the SED model in Section \ref{sec:backbone} as the backbone model.
        The backbone model consists of two layers of GAT, where each node in the batch aggregates messages from 800 direct neighbors and 100 one-hop neighbors.
        We set the mini-batch size to 2000, the learning rate to be $1e-3$, the margin for the contrastive triplet loss to 3, and employed the Adam optimizer.
        During the FL process, we perform 50 communication rounds, and each client conducts local training for 1 epoch, a compromise value across all baselines \cite{zhang2023fedala,chen2022personalized}.

        The baseline methods are implemented based on open-source implementations on Github\footnote{\url{https://github.com/dawenzi098/SFL-Structural-Federated-Learning}}\footnote{\url{https://github.com/zs2847037826zs/PFL-Non-IID/tree/master}}.
        For SFL, the client-wise relation graph is constructed based on the distances between model parameters using Euclidean distance.
        It is observed that FedALA failed to converge and entered a state of deadlock after the initial communication round. 
        Therefore, we set a local patience of 10 for FedALA.
        We set the number of epochs for local training to 50.

        For the robustness analysis, we conducted injection attacks involving model poisoning\footnote{\url{https://github.com/RingBDStack/KPGNN}} and data poisoning\footnote{\url{https://github.com/thunlp/HiddenKiller}} following \cite{zhang2022fldetector} and \cite{qi2021onion}, respectively. 
        % For the robustness analysis, we conducted injection attacks involving model poisoning and data poisoning following \cite{zhang2022fldetector} and \cite{qi2021onion}, respectively. 
        % The poison rate of the data poisoning attack is set to 0.2.
        % These attacks are conducted with the threat model outlined in Section \ref{sec:threat_model}, assuming the presence of a single malicious client.
        
        All methods utilize k-means clustering.
        All experiments are repeated 5 times to mitigate the uncertainty of deep learning methods.
        We report the average value and standard deviation of the 5 repetitions.
        All implementations are available at \url{https://github.com/XiaoyanWork/DAMe}.
        % All implementations are available on Github\footnote{\url{https://anonymous.4open.science/r/DAMe-284F/}}.

    \subsection{Evaluation Metric}
        Technically, SED involves learning representations of social messages and clustering them into specific events.  
        We evaluate the performance of all methods using three commonly used metrics for clustering tasks: Normalized Mutual Information (\textbf{NMI}) \cite{estevez2009normalized}, Adjusted Mutual Information (\textbf{AMI}) \cite{vinh2009information}, and Adjusted Rand Index (\textbf{ARI}) \cite{vinh2009information}.
        These metrics quantify the similarity between the detected and ground truth clusters. 
        A higher score of these metrics indicates better message representation.
\section{Experimental Results}

        \begin{figure*}[htbp]
            \centering
            \includegraphics[width=1.0\linewidth]{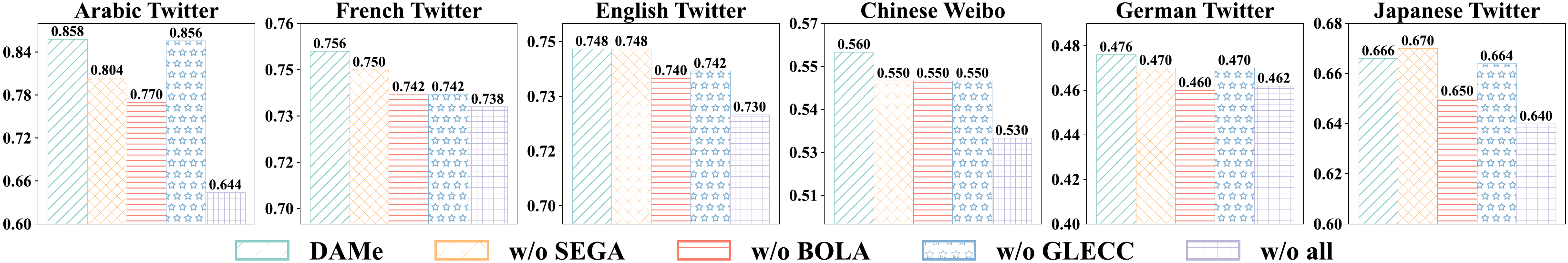}
            \caption{The results of ablation study on all datasets.}
            \label{fig:ablation}
        \end{figure*}

    \subsection{RQ1: Federated Performance}
        This section investigates the performance of SED with various FL systems.
        The result is demonstrated in Table \ref{tab:rq1}.
        The results show that DAMe has outperformed all baseline methods on all metrics for each client dataset.
        Especially on the Arabic dataset, which suffers from limited data samples, DAMe surpasses the baseline methods by at least 0.16, 0.17, and 0.16 regarding NMI, AMI, and ARI, respectively.
        This observation indicates that DAMe has the potential to integrate most external knowledge from other clients, thereby promoting local performance to the greatest extent.
        It is observed that compared to local training, DAMe achieves an average gain of 0.07, 0.09, and 0.14 in NMI, AMI, and ARI, respectively, surpassing all FL baseline methods. 
        These results highlight that the proposed framework meets the objective of FedSED, which is to enhance the performance of individual clients and benefit all clients involved.
        Moreover, in comparison to the pFL methods, the results reveal the following:
            (1) When comparing methods that directly override the local model with the global model (SFL, PerFedAvg, Ditto), these methods encounter challenges in achieving satisfactory local performance. 
            This is because, in each communication round, training is perceived as starting from scratch on the client side, initialized with the global parameters.
            In the case of PerFedAvg, fine-tuning on the client side with local data necessitates an additional training round after the FL process. 
            However, during local fine-tuning, there is a risk of catastrophic forgetting, whereby the learned information from the global model can be lost.
            As for SFL, which aggregates the global model using structural information on the server side, does not significantly contribute to local performance improvement. 
            This is because the objective is to personalize the model with local data, and a model that integrates the majority of external knowledge and overrides the local model disregards the importance of local characteristics, which are crucial factors in personalization. 
            Consequently, SFL does not surpass local training in most cases.
            (2) When comparing DAMe with methods that locally aggregate models (FedALA and APPLE), 
            it is observed that, for APPLE, exposing each client to other clients' local models does not necessarily lead to improved local performance. 
            This is because clients cannot accurately determine which models are useful or possess similar distributions to their own. 
            As for FedALA, the performance is relatively satisfactory due to its parameter-level aggregation, which aids in identifying relevant knowledge within the global model. 
            However, FedALA's approach of dispatching the global model based on weighted averages of data samples does not provide clients with the most advantageous information they require. 
            In contrast, our proposed framework addresses this limitation by considering client similarity during the global aggregation process. 
            This allows the server to dispatch an aggregated global model better aligned with each client's specific needs.
        To sum up, the proposed dual aggregation mechanism significantly enhances the performance of local training. 
        This is achieved by considering the local characteristics and providing clients with the knowledge they require the most. 
        The dual aggregation mechanism is crucial in improving local performance to a paramount extent.

        \begin{table}[t]
        \aboverulesep=0ex
        \belowrulesep=0ex
        \caption{Results of injection attack towards DAMe.} 
        \centering
        \resizebox{1.0\linewidth}{!}{
        \begin{tabular}{l|c|c|c}
        \toprule
                         & Clean         & Model Poisoning & Data Poisoning \\ 
        \hline
        % English Twitter  & .76 $\pm$ .00 & .70 $\pm$ .00   & .67 $\pm$ .00  \\
        French Twitter   & .76 $\pm$ .00 & .75 $\pm$ .00   & .75 $\pm$ .00  \\
        Arabic Twitter   & .86 $\pm$ .00 & .85 $\pm$ .07   & .85 $\pm$ .00  \\
        Japanese Twitter & .67 $\pm$ .00 & .67 $\pm$ .00   & .67 $\pm$ .00  \\
        German Twitter   & .48 $\pm$ .00 & .47 $\pm$ .00   & .45 $\pm$ .00  \\
        Chinese Weibo    & .56 $\pm$ .00 & .55 $\pm$ .02   & .55 $\pm$ .00  \\
        \hline
        % Average          & .69           & .67             & .66            \\ 
        Average          & .69           & .66             & .65            \\ 
        \bottomrule
        \end{tabular}
        }
        \label{tab:rq2}
        \end{table}

    \subsection{RQ2: Robustness Analysis}\label{sec:robust}
        % In this section, we conduct a robustness analysis to evaluate how well DAMe can handle the injection attack in a federated setting.
        To evaluate DAMe's robustness, we investigate its resistance to two types of attacks: model poisoning attacks and data poisoning attacks.
        In both attacks, the malicious client possesses the English dataset and aims to compromise the FL process. 
        In the model poisoning attack, the client uploads manipulated parameters to the server \cite{zhang2022fldetector}, resulting in a model with no practical utility. 
        In the data poisoning attack, the client injects backdoors into its own data \cite{qi2021onion} and trains a local model using the poisoned data. 
        Consequently, the injected backdoor affects the model parameters and corrupts the global model.
        In traditional settings, the server integrates all received models, including the poisoned model, without differentiation, which decreases the accuracy of the global model. 
        Moreover, this corrupted global model is dispatched to individual clients, overriding their local models and compromising their local training.
        However, the proposed dual mechanism has advantages against such attacks.
        With BOLA, clients can determine the proportion of the dispatched global model that should be integrated into their local training. 
        This enables them to avoid incorporating potentially poisoned parameters. 
        Additionally, SEGA, implemented on the server side, leverages the client-uploaded models to compute the similarity between clients' models. 
        This similarity calculation enables the server to develop a dispatching strategy, ensuring that models containing poison parameters are not distributed to clients.
        % By employing these mechanisms, our DAMe framework enhances robustness against injection attacks in an FL scenario.
        From the result in Table \ref{tab:rq2}, it is evident that the performance of the DAMe on all datasets demonstrates a negligible perturbation in performance. 
        This observation indicates that DAMe is resilient to injection attacks in FL.
        % , as its performance is not affected by them.
        Regarding injection attacks, DAMe has the capability to disregard model injection, as it selectively focuses on relevant information that is essential for clients (while disregarding irrelevant noise).

        \begin{figure}[b]
            \centering
            \includegraphics[width=1.0\linewidth]{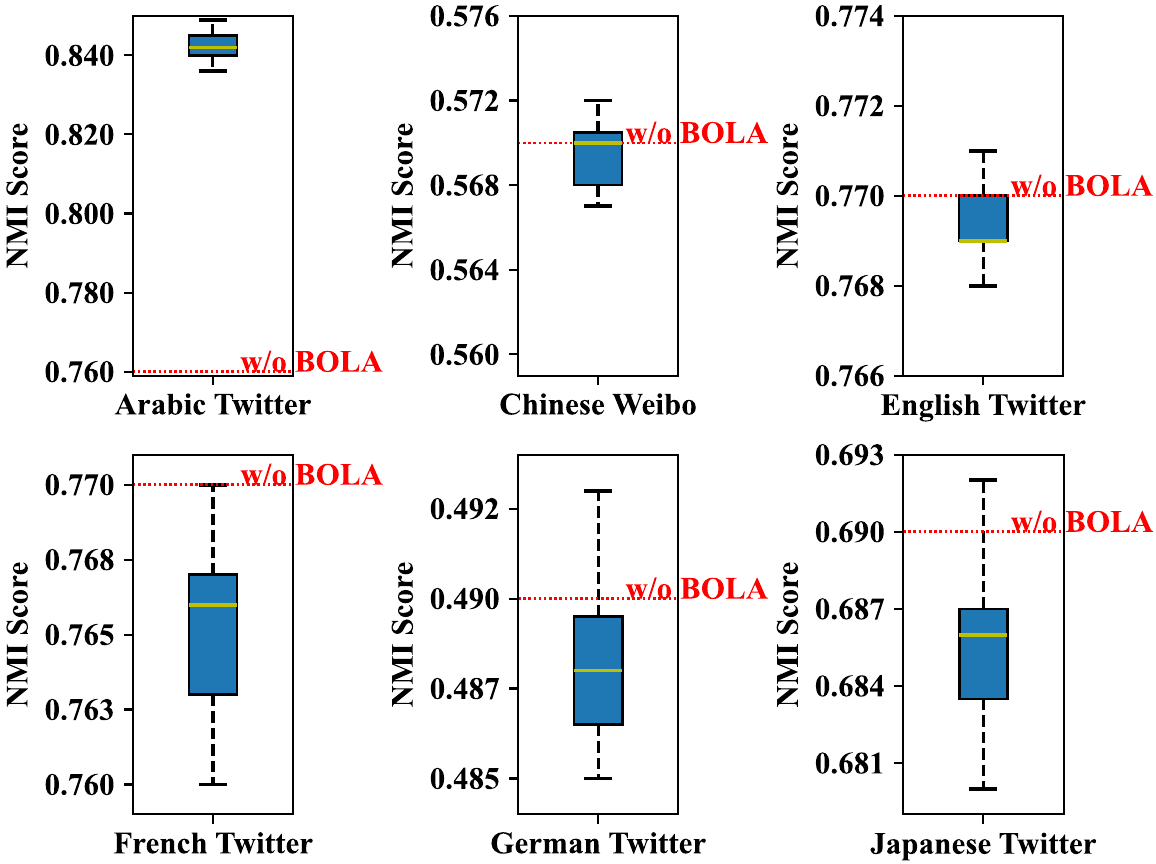}
            \caption{
            The NMI score corresponds to the aggregation weight within the BOLA search space. The overall search space is delineated by the two black lines, while the blue box represents 50\% of the NMI scores associated with the search space, and the yellow line denotes the midpoint. The red dotted horizontal line illustrates the performance of DAMe without BOLA.
            }
            \label{fig:bayesian}
        \end{figure}

        \begin{figure*}[htbp]
            \centering
            \includegraphics[width=1.0\linewidth]{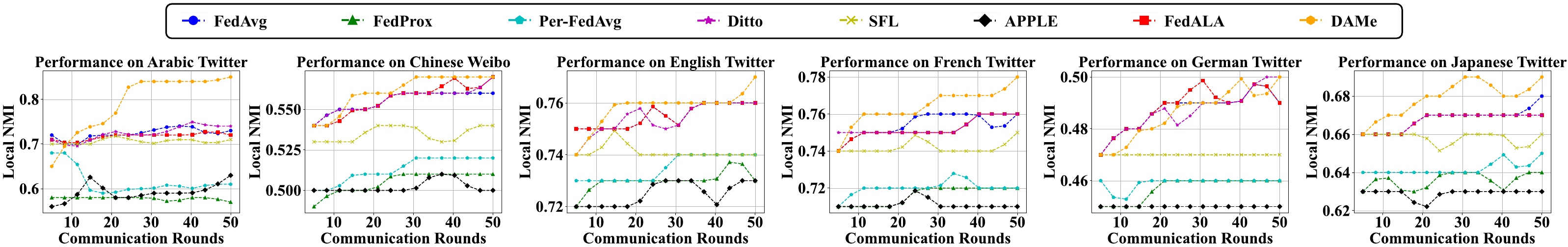}
            \caption{The convergence plots of all methods.}
            \label{fig:converge}
        \end{figure*}

    \subsection{RQ3: Analysis on DAMe}
        This section presents an ablation study to identify the contribution of the components in DAMe and analyze the effect of the modules.

        The results of the ablation study are presented in Figure \ref{fig:ablation}. 
        The findings demonstrate that all proposed components enhance the performance of FedSED.
        Specifically, BOLA exhibits a substantial influence on performance improvement. 
        This indicates that in pFL settings, it is crucial to empower clients to determine how much they incorporate knowledge from the global model, while preserving their individual characteristics during training.
        Moreover, SEGA significantly improves performance and surpasses the baseline SFL, aggregating the global model based on structural information within a client graph.
        This finding suggests that our approach, which leverages client similarity and minimizes the 2DSE of the client graph for global aggregation, effectively identifies pertinent knowledge that boosts local performance.
        Lastly, GLECC also plays a role in improving overall performance.
        This implies that aligning the global and local representations of the same event is advantageous for achieving a consensus between the global and local models, thereby mitigating the inherent heterogeneity.

        % Figure \ref{fig:bayesian} presents the NMI scores of DAMe without BOLA (red line) and the NMI scores associated with the Bayesian aggregation weight search space in DAMe (blue rectangle).
        We analyze the Bayesian search space in the last communication round of the FL process and present the visualization of BOLA in Figure \ref{fig:bayesian} .
        For all clients, DAMe achieves the best performance.
        For French, German, and Japanese Twitter, DAMe without BOLA achieves performance above the first quartile of the weight distribution in the search space of the Bayesian optimization algorithm but consistently lags behind the upper bound.
        This observation highlights the significant benefits of utilizing BOLA to search for local aggregation weights, thereby enhancing local performance.

    \subsection{RQ4: Overhead Analysis}
        Here, we delve into the crucial factors in FL, including the convergence, computation and communication overhead of the proposed framework and baseline methods.

        \textbf{Convergence}
        As depicted in Figure \ref{fig:converge}, the proposed DAMe demonstrates stable convergence. 
        This stability can be attributed to DAMe's tailored design for the task of SED, which focuses on learning message representations and performing clustering.
        DAMe stands out as the first work to consider the characteristics of the SED task, unlike baseline methods, which serve as a general framework for FL.
        This task-specific approach allows DAMe to better suit the requirements of SED, enhancing its performance and convergence compared to the less specialized FL baselines.

        % \begin{figure*}[htbp]
        %     \centering
        %     \includegraphics[width=1.0\linewidth]{Img/plot.pdf}
        %     \caption{The convergence plots of all methods.}
        %     \label{fig:converge}
        % \end{figure*}
        
        \textbf{Computation Overhead}
        The second column in Table \ref{tab:rq4} presents the time consumption of all methods during the entire training process.
        It is observed that DAMe without the GLECC module has the shortest time for training in the FL setting, shorter than DAMe, since it omit the process of obtaining the global and local event representation. 
        % This can be attributed to the computational requirements of the GLECC module, which involves calculating the triplet loss for both the global and aggregated models on local data, computing event representations within each batch, and evaluating pairwise losses between event representations. 
        Moreover, Per-FedAvg and Ditto exhibit the longest training times, surpassing DAMe's duration by 1.5 times. 
        This can be attributed to their approaches of local fine-tuning or training additional local models, which significantly prolongs the training process. 
        However, contrasting the results in Table \ref{tab:rq1}, these methods do not demonstrate substantial performance improvements. 
        % This indicates a suboptimal trade-off between computation and performance, as the additional time spent on training does not yield significant enhancements in performance.
        % Additionally, FedALA faces a challenging convergence issue during training, particularly after the first communication round. 
        % The difficulty arises from the struggle to determine the optimal merging weight for all parameters in the model. 
        % It is worth noting that the parameter scale of the SED task model is larger than that mentioned in their original work.
        % Considering the diverse nature of FL tasks, FL frameworks should be capable of handling variations in parameter scale. 
        In this regard, DAMe demonstrates success in such scenarios by achieving notable task performance while maintaining reasonable computational overhead.

        \textbf{Communication Overhead}
        The communication overhead is shown in the last column in Table \ref{tab:rq4}.
        In most scenarios, the methods adhere to the centralized FL setting, where a single server communicates with all clients, leading to a consistent communication overhead across the same parameter scales.
        However, APPLE adopts a decentralized FL setting, where all clients interact with each other to determine a local aggregation strategy. 
        This significantly increases the communication overhead as the number of clients grows.
        In our experiment, where $K=6$, the communication overhead of APPLE is 3.5 times higher compared to other methods.

        \begin{table}[htbp]
        \aboverulesep=0ex
        \belowrulesep=0ex
        \caption{Computation and communication overhead of all methods with 50 communication rounds and 1 local epoch. $\Sigma$ denotes the scale of model parameters, $K$ represent the number of clients. For local training, we report the sum of training time across all datasets.} 
        \centering
        \resizebox{1.0\linewidth}{!}{
        \begin{tabular}{l|cc|c}
        \toprule
                                       & \multicolumn{2}{c|}{Computation} & Communication    \\
        \multicolumn{1}{c|}{}          & (Total time)    & (Time/round)   & (Param./round)   \\
        \hline
        Local                          & 1402 min        & -              & -                \\ 
        \hline
        FedAvg                         & 1522 min        & 30 min         & $2 * \Sigma$     \\
        FedProx                        & 1902 min        & 38 min         & $2 * \Sigma$     \\ 
        \hline
        Per-FedAvg                     & 2945 min        & 59 min         & $2 * \Sigma$     \\
        Ditto                          & 2890 min        & 58 min         & $2 * \Sigma$     \\
        SFL                            & 1555 min        & 31 min         & $2 * \Sigma$     \\
        APPLE                          & 1568 min        & 31 min         & $(1+K) * \Sigma$ \\
        FedALA                         & 2725 min        & 55 min         & $2 * \Sigma$     \\ 
        \hline
        DAMe                           & 1896 min        & 38 min         & $2 * \Sigma$     \\
        \multicolumn{1}{r|}{w/o BOLA}  & 1688 min        & 33 min         & $2 * \Sigma$     \\
        \multicolumn{1}{r|}{w/o SEGA}  & 1612 min        & 32 min         & $2 * \Sigma$     \\
        \multicolumn{1}{r|}{w/o GLECC} & 1530 min        & 31 min         & $2 * \Sigma$     \\ 
        \bottomrule
        \end{tabular}
        }
        \label{tab:rq4}
        \end{table}
\section{Conclusion}
    In this paper, we initiate the study on FedSED and propose DAMe, a personalized federated framework for social event detection that incorporates two aggregation strategies.
    In DAMe, the server provides clients with maximum external knowledge with a structural entropy-based global aggregation strategy; clients leverage received knowledge and retain local characteristics to the greatest extent by a Bayesian optimization-based local aggregation strategy.
    Moreover, the local optimization process is guided by an event-centric constraint that mitigates the issues arising from heterogeneity, while preventing overfitting to the local data.
    Extensive experiments on six SED datasets across six languages and two platforms have demonstrated the effectiveness of DAMe.
    Further robustness analyses have shown that DAMe is resistant to federated injection attacks.

\begin{acks}
This work is supported by the National Key Research and Development Program of China (No.2023YFF0905302),  and the Yunnan Provincial Major Science and Technology Special Plan Projects (No.202302AD080003).
% the National Natural Science Foundation of China (No.62322202),
\end{acks}

\bibliographystyle{ACM-Reference-Format}
\bibliography{ref}


\end{document}